%% file: fatml-paper.tex
\newif\ifnotes
\notesfalse

\newif\ifplain
\plainfalse

\newif\ifacm
\acmtrue

\newif\ifpmlr
\ifplain
  \pmlrfalse
\else
  \ifacm
  \pmlrfalse
  \else
  \pmlrtrue
  \fi
\fi

\PassOptionsToPackage{hyphens}{url}
\ifplain
  \documentclass[11pt]{article}
  \usepackage{xcolor}
  \usepackage{fullpage}
  \usepackage{times}
  \usepackage{relsize}
  \usepackage{setspace}
  \usepackage[backend=bibtex,citestyle=verbose-ibid]{biblatex}
  \renewcommand{\cite}[1]{\footcite{#1}}
  \usepackage{hyperref}
\else
\ifacm
  \documentclass[sigconf, anonymous]{acmart}
  \usepackage{xcolor}
  \definecolor{linkcolor}{rgb}{0 0.8 0}
  \AtEndPreamble{\hypersetup{pdfstartview=FitH,hyperindex=true,pdfpagemode=UseOutlines,bookmarksnumbered=true,bookmarksopen=true,bookmarksopenlevel=2,
      colorlinks=true,hidelinks=false,pdfborder={0 0 0.8},linkbordercolor=linkcolor,citebordercolor=linkcolor,urlbordercolor=linkcolor}}
  \usepackage{balance}
\else
  \documentclass[pmlr,twocolumn,10pt]{jmlr}
  \renewcommand{\cite}[1]{\citep{#1}}
\fi
\fi
\usepackage{booktabs} 

\ifnotes
  \newcommand{\colornote}[3]{\textcolor{#1}{\smaller\sffamily\bfseries\boldmath[\ignorespaces #2 --- \smaller\textit{#3}]}}
\else
  \newcommand{\colornote}[3]{}
\fi

\ifplain
\else
\ifacm
\setcopyright{rightsretained}

  \acmConference[FAT* 2018]{FAT* conference}{February 2018}{New York, NY USA} 
  \acmYear{2018}
  \copyrightyear{2018}

\else
  \jmlrvolume{81}
  \jmlryear{2018}
  \jmlrworkshop{Conference on Fairness, Accountability, and Transparency}
  \editors{Sorelle A. Friedler and Christo Wilson}
\fi
\fi

\title{Interventions over Predictions: \\ Reframing the Ethical Debate for Actuarial Risk Assessment}

\ifplain
\else
\ifacm
\author{Chelsea Barabas}
\affiliation{%
  \institution{MIT Media Lab}
}
\email{cbarabas@media.mit.edu}

\author{Karthik Dinakar}
\affiliation{%
  \institution{MIT Media Lab}
  }
\email{karthik@media.mit.edu}

\author{Joichi Ito}
\affiliation{%
  \institution{MIT Media Lab}
}
\email{joi@media.mit.edu}

\author{Madars Virza}
\affiliation{%
  \institution{MIT Media Lab}
}
\email{madars@mit.edu}

\author{Jonathan Zittrain}
\affiliation{%
  \institution{Harvard University}
}
\email{a2jz@law.harvard.edu }

\renewcommand{\shortauthors}{Barabas et~al.}

\begin{abstract}
\input{abstract}
\end{abstract}
\else
\author{\Name{Chelsea Barabas} \Email{cbarabas@media.mit.edu}\\
  \Name{Karthik Dinakar} \Email{karthik@media.mit.edu} \\
  \Name{Joichi Ito} \Email{joi@media.mit.edu} \\
  \Name{Madars Virza} \Email{madars@mit.edu} \\
  \addr MIT Media Lab, 75 Amherst St. E15-245, Cambridge, MA 02139, United States
\AND
  \Name{Jonathan Zittrain} \Email{a2jz@law.harvard.edu}\\
  \addr Harvard University, Griswold 505, 1525 Massachusetts Avenue, Cambridge, MA 02138, United States
}
\fi
\fi

\begin{document}

\maketitle
\ifplain
  \begin{abstract}
  \input{abstract}
  \end{abstract}
\else
\ifacm
\else
  \begin{abstract}
  \input{abstract}
  \end{abstract}
\fi
\fi
%
%

\ifplain
\else
\ifacm
\else
\fi
\fi

\input{introduction}
\input{current-debates}
\input{predictive-or-diagnostic}
\input{regression-ml-ci}
\input{towards-causal}
\input{conclusion}

\ifplain
  \section*{Acknowledgements}
  \label{sec:acks}
  \input{acks}
\else
  \ifacm
  %
\begin{acks}
\input{acks}
\end{acks}
  \else
  \acks{\input{acks}}
  \fi
\fi

\ifplain
\else
\ifacm
  \balance
  \bibliographystyle{ACM-Reference-Format}
  \bibliography{bibliography}
\else
  \bibliography{bibliography}
\fi
\fi

\end{document}

%% file: abstract.tex
Actuarial risk assessments might be unduly perceived as a neutral way to counteract implicit bias and increase the fairness of decisions made at almost every juncture of the criminal justice system, from pretrial release to sentencing, parole and probation. In recent times these assessments have come under increased scrutiny, as critics claim that the statistical techniques underlying them might reproduce existing patterns of discrimination and historical biases that are reflected in the data. Much of this debate is centered around competing notions of fairness and predictive accuracy, resting on the contested use of variables that act as ``proxies'' for characteristics legally protected against discrimination, such as race and gender.

We argue that a core ethical debate surrounding the use of regression in risk assessments is not simply one of bias or accuracy. Rather, it's one of purpose. If machine learning is operationalized merely in the service of predicting individual future crime, then it becomes difficult to break cycles of criminalization that are driven by the iatrogenic effects of the criminal justice system itself.  We posit that machine learning should not be used for prediction, but rather to surface covariates that are fed into a causal model for understanding the social, structural and psychological drivers of crime. We propose an alternative application of machine learning and causal inference away from predicting risk scores to risk mitigation.

%% file: introduction.tex
\section{Introduction}
\label{sec:introduction}
In 2016, a team of investigative journalists from ProPublica published an article \cite{AngwinLMK16} claiming that COMPAS, a proprietary risk assessment tool that has been used in the U.S. criminal justice system, was racially biased. The article sparked a national debate about the expanding use of algorithmic decision-making aids in the courts.  This debate is emblematic of a much broader conversation around the appropriate use of data and statistical methods in society, in spheres as varied as health care, housing, employment and education.
%
%
For example, Selbst \cite{Selbst16} cites media debates on consumer finance \cite{Economist17}, employment \cite{Wang17}, housing \cite{Biggs16}, health care \cite{Siwicki17}, and sentencing \cite{Tashea17}.
While these issues are largely framed in terms of new technology, driven by ``big data'' or ``artificial intelligence,'' the methods and tools in question are often incremental iterations on much older actuarial decision-making practices. This is certainly the case for risk assessments, which have existed in 
some form in the U.S. criminal justice system since the 1920's.
%
%

By placing the current debate around risk assessment in a broader context, we can get a fuller understanding of the way these actuarial tools have evolved to achieve a varied set of social and institutional agendas. The current debate around risk assessment tends to characterize or implicitly concede the purpose of these tools as fundamentally predictive in nature. The more significant power of data-driven risk assessment can be as a broader diagnostic tool, one used to help practitioners address risk as a dynamic, intervenable phenomenon. When risk assessments are recast in this light, we can ask whether or not regression and machine learning methods can help in diagnosis and intervention, rather than prediction. We make a case for why causal inference, a statistical method designed explicitly to establish a causal connection between a ``treatment'' and an outcome of interest, provides a more desirable framework for developing intervention-driven risk assessments in the criminal justice system.

%% file: current-debates.tex
\section{Current debates on the fairness of risk assessments}
\label{sec:current-debates}
Risk assessment tools have been adopted to assist with a number of decision points throughout the criminal justice system, from pretrial release to post-conviction sentencing, probation and parole. In spite of the tools' growing influence over judicial or administrative decisions, defendants rarely have an opportunity to probe or challenge the recommendations the tools generate. While defendants are supplied with their end scores, they are typically not entitled to access the calculations or input data that were used to calculate the final tabulation. Critics have called for increased transparency surrounding the development and administration of these tools, arguing that the proprietary interests of assessment developers should not take precedence over a defendant's right to due process in the courts \cite{Wexler17,AngwinLMK16,FennelH80}. In addition to transparency issues related to proprietary software, a major focus of recent scholarship has been on developing interpretable machine learning models that enable outside researchers to better scrutinize the underlying logic of actuarial tools which are based on machine learning \cite{ZengUR17,VellidoMGL12}.
%
%
%
%

Others have focused on the particular ethical challenges that arise during the administration and interpretation of risk assessments. Although risk assessments can be perceived -- and sometimes marketed -- as an objective means of overcoming human bias in decision making, there remain many ways that human discretion enters into the formation and interpretation of risk scores. Ethnographic work has shown that scores are frequently misapplied and misinterpreted by judges and practitioners \cite{HannahMoffat15}.

Moreover, scholars have documented specific strategies that administrators can use to shade or manipulate end scores to reflect their preconceptions of offender risk \cite{HannahMoffat15,Hardy14}. There have also been multiple documented instances in which a risk or needs score has been inappropriately referenced for a decision it was not intended to inform (for example, a needs score that was intended to inform probation decisions being cited in judges' sentencing statements) \cite{Wexler17,AngwinLMK16}. At least one recent court ruling has emphasized the importance of providing judges additional information about the limitations inherent in risk score predictions \cite{Loomis15}. Others have called for more work to be done in order to understand how risk scores interact with the pre-existing biases and beliefs of the individuals who rely on them \cite{FennelH80}.
%
%

Amidst all these issues, arguably the primary focus of debate regarding the fairness of risk assessments has centered around the specific methods and data used to develop statistically valid predictions. Risk assessment is commonly characterized as a predictive technology, one whose value is measured by the degree to which they are more accurate than the fallible human decision makers they are designed to advise or supplant \cite{KleinbergLLLM17}. This characterization stems in part from the fact that much of the recent attention on risk assessments has centered around their appropriate use in pretrial release decisions \cite{AngwinLMK16,KleinbergLLLM17,FloresBL16}.

In contrast to other stages of the criminal justice process, pretrial release is framed as a narrow and fundamentally predictive task: risk assessments are used to inform decisions about whom to detain and whom to release before a trial date. Increasingly, pretrial risk assessments are being embraced as part of a larger effort to move away from a ``resource-based'' approach of making release decisions, based on individuals' ability to post cash bail, to a ``risk-based'' approach that centers on predicting one's likelihood of engaging in a specific set of undesirable behaviors, such as failure to appear in court or to engage in new criminal activity \cite{arnoldF,pji}.

In this context, some scholars have framed the value-add of risk assessment strictly in terms of their ability to make predictions more accurately than the judges meting out such decisions by themselves \cite{KleinbergLLLM17}. Because the value-add of risk assessments has been framed largely in terms of their predictive power, much of the ethical debate surrounding them has focused on how to handle various forms of bias in the face of less-than-accurate tools. Scholars have identified numerous ways that statistically-driven methods like machine learning can reproduce existing patterns of individual prejudice and institutionalized bias. These mechanisms correspond with different stages in the model design process such as identifying the research question, collecting training data, labeling examples within the training set, feature selection and double-encoding bias in proxies for protected classes \cite{BarocasS14,Gong16,AngwinLMK16,Starr14,Citron16}.

Others have focused specifically on bias in model outcomes. Critics of risk assessment have expressed concerns about the distribution of inter-group inaccuracy in risk models. They emphasize the importance of predictive parity as an explicit goal; that is, the systems we use should not only be equally accurate, but also have similar accuracy rates over all test groups (e.g. different racial groups or genders) to which they are applied. These researchers have argued that risk assessments are not that accurate to begin with, and for the large percentage of outcomes that is not accurate, disparate impact tends to fall disproportionately across the population, along racial lines \cite{AngwinLMK16,Citron16}.

Yet proponents of risk assessment have argued that uneven distributions of false positives and false negatives do not constitute a ``biased'' model in a strictly technical sense. These individuals contend that the most important consideration regarding bias is whether or not the probability estimates provided by the algorithm are well-calibrated, so that the algorithm performs in a way that is equally accurate across all groups \cite{FloresBL16,SkeemML16}

This debate has led to a broader discussion on the inherent trade-offs in these competing notions of fairness. Kleinberg et al. \cite{KleinbergMR17} argue that, in general, no mechanism can both achieve optimal accuracy, and optimal predictive parity, unless the base rates for the groups are equal, or the algorithm in fact achieves perfect classification. In their excellent survey Berk. et al:  identify six kinds of fairness, and show that not only these notions conflict with accuracy, but also with one another \cite{BerkHJKR17}.

This debate is now evolving to identify specific trade-offs between things like predictive accuracy and predictive parity across groups. This makes sense when we frame the value of risk assessments strictly in terms of their utility as a predictive technology, whereby accuracy levels are often in direct tension with these competing notions of fairness. However, this framing of the issues eschews the fact that risk assessments are rarely used just for predictive purposes. The majority of risk assessments in use today are based on a ``risk-needs-responsivity'' framework to inform a broad range of decisions regarding effective punishment and treatment \cite{hannah2005criminogenic}. Such assessments are, and should be, fundamentally diagnostic in nature, whereby risk is conceptualized as a dynamic phenomenon that can be lowered through a range of semi-personalized interventions. 

In the criminal justice system, risk assessments play an integral role in shaping how practitioners understand and intervene in the lives of offenders in the service of lowering future crime rates. We conceptualize risk assessments as part of an assemblage of practices and tools that give rise to visibilities, or ways of viewing, individual offenders and complex phenomena like criminal behavior \cite{Hardy14}. Risk assessments have evolved over the last several decades to support a diverse set of institutional goals and processes. In the following section, we give a  brief overview of how risk assessments have evolved from a tool used solely for prediction to one that is diagnostic at its core. We then call into question whether or not the statistical methods currently underlying risk assessments, namely regression and, to a limited extent, machine learning, are the appropriate methods to use in the service of this goal.

%% file: predictive-or-diagnostic.tex
\section{Risk assessments: predictive or diagnostic tools?}
\label{sec:predictive-or-diagnostic}
The evolution of actuarial risk assessment has been well documented by others, and is frequently broken down into four generations of tools \cite{AndrewsBW06,Harcourt15,MonahanS16,DesmaraisS13,HannahMoffat13}. The first generation of risk assessments emerged in the 1920's and were based largely on semi-structured clinical evaluations that were carried out by skilled professionals as part of an effort to identify rehabilitative treatment options for offenders. These assessments were structured around a standardized set of clinical items, but did not include a statistical mechanism for validating scores and decisions \cite{AndrewsBW06}.

First generation risk assessment tools fell out of fashion in the 1970's, as they were widely criticized for being too subjective and having low levels of predictive accuracy \cite{HannahMoffat13}. Also during this time, a ``nothing works'' mindset began to supplant the rehabilitative approach that had dominated the correctional system during the first half of the twentieth century \cite{MauruttoHM06}. In place of treatment programs, approaches based on incapacitation and other ``law and order'' measures became the norm.

A second generation of assessment was optimized and validated for predictive accuracy using a new statistical method: regression modeling. Regression modeling is particularly well-suited for prediction-oriented assessment, because it enables researchers to identify variables that are predictive of an outcome of interest, without necessarily having to understand why that factor is significant. To this end, the second generation of assessment focused almost exclusively on static historical factors, such as age and criminal history. Such factors were considered desirable because they were perceived as both objective and highly predictive \cite{MauruttoHM06}. This marked a subtle shift in the values and goals risk assessments were designed to support, away from semi-personalized treatment and towards objective prediction. As a result, risk assessments were reconfigured as a more simplistic tool that relied heavily on prior criminal history to predict future outcomes.

These changes marked a significant shift in the way the justice system understood and responded to the notion of offender risk. While earlier tools were designed to facilitate effective treatment, new statistical assessments were based on a static notion of risk, one that was not easily changed through intervention. These assessments gave rise to what some scholars have dubbed a ``new penology,'' whereby penal policies shifted away from rehabilitative interventions to more administrative approaches to population management that rely almost exclusively on incapacitation to mitigate risk \cite{FeeleyS92}. As Maurutto and Hannah-Moffat argue, when assessments pivot towards prediction, rather than intervention, ``risk is conceived of as a negative strategy that incapacitates and manages but never produces productive transformations'' \cite{MauruttoHM06}. Since the 1970s, the shift towards standardized decision-making and away from individualized treatment accompanied a sharp increase in incarceration rates \cite{harcourt2010risk}.

These prediction-oriented risk assessments were a part of a suite of tools and policies introduced in the 1970's and 80's that relied heavily on prior criminal history to predict and systematize interventions to prevent future crime. Other policies included state and federal sentencing guidelines, which were ostensibly designed to standardize decision making across jurisdictions and minimize bias in sentencing. A growing body of scholarship has linked these ``colorblind'' policies to trends in mass incarceration and growing racial disparities in the criminal justice system. \cite{VanCleveM15,Schlesinger11,Harcourt15,Hamilton15BackToFuture}.
%
%
%
%

Scholars have argued that prediction-oriented risk assessments produce a ``ratchet effect'' on profiled populations, because they fail to take into consideration the criminogenic nature of the criminal justice system itself \cite{Harcourt15,Hamilton15BackToFuture}. Studies have demonstrated diminishing returns on the use of incarceration as a means of lowering crime \cite{Zimring06}. In fact, the ripple effects of incarceration, such as the weakening of family ties and diminishing employment opportunities, can exacerbate the likelihood of future recidivism at both the individual and collective level \cite{CullenJN11,DeFinaH10,MitchellCMB17}. As Hamilton has argued, assessments that rely on a static, prediction-oriented notion of risk can act as a ``self-fulfilling prophecy'' that justifies and widens the net of social control over marginalized populations \cite{Hamilton15BackToFuture}. When risk is constituted as a static phenomenon, it can only suggest what should not be done (i.e. release), and gives little insight into what can be done to improve outcomes \cite{Beck92}.

By the 1980's, critiques of this prediction-oriented approach started to emerge, heralding in the next iteration of assessment. Proponents of evidence-based justice reform argued that risk assessment should be reconfigured to help in the effort of reducing the risk of recidivism and lowering incarceration rates. Risk assessments were reframed as a way of effectively identifying and allotting scarce resources to programs and interventions that were ``proven'' to work
\cite{AndrewsB10,AndrewsBH90,CullenG01,Etienne09},

To this end, a third generation of actuarial tools integrated factors that were conceived as ``criminogenic needs,'' or intervenable factors that are believed to impact one's risk of recidivating. These included a more diverse set of inputs, such as employment status and history of substance abuse, which effectively reconstituted risk as a dynamic phenomenon that could be intervened upon \cite{AndrewsBW06}. While many of these dynamic factors held less predictive power than the static attributes included on second generation tools, these needs factors were desirable because they enabled criminal justice practitioners to consider a broader set of treatment options aimed at lowering risk.

Thus, predictive risk and intervenable needs factors were combined into a hybrid risk/needs model, one that used regression to identify statistically significant risk/needs factors. The key difference between second and third generation assessments, then, was the incorporation of these less predictive, but dynamic, risk/needs factors that could inform the selection of treatment interventions beyond incapacitation. By the 1990's, these efforts were further supplemented by fourth generation tools that added ``responsivity factors,'' such as levels of intelligence, self-esteem, and psychological disorders, in order to improve response outcomes to treatment \cite{AndrewsB10}. The risk-needs-responsivity (RNR) framework has reasserted the authority of a correctional treatment approach, one that envisions offenders as individuals capable of reform through intervention.

Today risk assessments are used for two primary purposes, which Monahan and Skeem deem ``prediction-oriented'' and ``reduction-oriented'' approaches to assessment \cite{MonahanS16}. Prediction-oriented assessments are used to facilitate accurate and efficient prediction of future recidivism, while reduction-oriented tools are intended to inform treatment and supervision plans. In recent years, we've seen a resurgence of public interest in prediction-oriented assessments, particularly at the pretrial stage where release decisions are viewed as essentially a forecasting task. Yet we argue that even pretrial decisions should be characterized as a moment of intervention, rather than mere prediction. The push to adopt pretrial assessments, like the Public Safety Assessment, is tightly coupled with efforts to eliminate cash bail, which empirical analysis has demonstrated to be ineffective at lowering near-term risks (failure to appear and new criminal activity) and long-term recidivism rates. For example, Gupta et.al show that high bail amounts, which often result in pretrial detention, drive a rise in long-term recidivism rates \cite{gupta2016heavy}. 
%
%

Bail is an ineffective pretrial intervention that proponents of risk assessment would like to eliminate in the service of achieving better pretrial outcomes. When framed in this way, the logical next question is not so much whether or not judges make more accurate decisions when they use risk assessments. Rather, we should be asking whether or not pretrial risk assessments effectively support judges in making better decisions about whom to release, and under what conditions (i.e. how should the criminal justice system intervene in an individual's life to mitigate specific, relevant risks). Thus, we argue that the most socially beneficial use of risk assessments should be centered around their utility as a diagnostic tool, at all stages of the criminal justice life cycle. This is an important point because it informs the goals and characteristics we strive to achieve in the development of a fair and ethical risk assessment tool.

In recent times, there has been a growing enthusiasm around the idea of using machine learning techniques to enhance the performance of modern day assessments. However, these methods run the risk of swinging the trend of assessment back towards prediction, rather than intervention. New statistical techniques like machine learning have garnered a lot of enthusiasm and, as Kleinberg et al. have pointed out, this increased enthusiasm and investment might unduly push institutions to reframe their goals in a way that is amenable to machine learning analysis \cite{KleinbergLMO15}.

Similar to regression, machine learning models are well-suited to the task of identifying correlational relationships across an even wider range of variables, with the goal being to create assessments that can predict risk with greater accuracy. While earlier assessments were based on theories in criminology, data mining enables the surfacing of new patterns and trends that are ``born from the data,'' without necessarily having a theoretical explanation for why they emerge \cite{Kitchin14}. As such, machine learning is touted as a way of helping us identify patterns that we could not have discovered by ourselves, in the service of creating more predictive tools.

Yet, diagnostic practices do not map cleanly to tools that use prediction-oriented methods, such as regression and machine learning. While the goals of risk assessment have expanded, the statistical methods used to identify risks and needs have remained the same since their introduction in second generation tools. This regression-based methodology is ill-suited for the purpose of effective diagnosis and intervention of criminogenic needs. When we use regression for intervention, we run the risk of conflating correlational variables with causal ones, which gives rise to a range of challenges.

In the following section, we argue that assessments designed for intervention require a different set of statistical techniques, ones that can identify causal relationships between risk factors and future crime. We outline key differences between regression, machine learning and causal inference in order to make the case for moving away from using regression and machine learning for intervention-oriented assessment.

%% file: regression-ml-ci.tex
\section{Regression vs. Machine Learning vs. Causal Inference}
\label{sec:regression-ml-ci}
Nothwithstanding the popular discourse on the ethical use of risk assessments, the vast majority of these tools do not use new statistical methods frequently associated with ``artificial intelligence,'' such as machine learning. They are overwhelmingly based on regression models. Moreover, the small number of tools that are currently being developed using machine learning methods should be characterized as an incremental, rather than a transformational, innovation in the way risk assessments have historically worked.

Regression analysis is widely used for purposes of forecasting future events. The main goal of regression is to identify a set of variables that are predictive of a given outcome variable. This is achieved by determining the optimal weights for a given set of covariates, ones that are best predictive of the outcome variable of interest. This is done through processes called model checking and selection \cite{gelman2007data}, whereby statistical tests are run on each covariate to see how significantly predictive they are of the outcome variable . Covariates that are identified through this process are correlational, not causal, of the outcome variable. For example, \textit{prior arrest} has been found to be an important predictor of future crime but does not shed light on the causal drivers of criminal behavior. Regression based models, while better suited for predicting risk scores, are not well-equipped to answer causal questions on interventional covariates. Tacking on interventional covariates to a list of risk covariates and using regression models for risk prediction suffers from three main drawbacks:

\emph{First}, regression models for risk prediction are blunt instruments that do not always contextualize risk and intervenable factors specific to different subgroups in the data \cite{Greiner08}. Critics have pointed out that criminogenic needs in risk assessments are treated as ``universal,'' even though the theories and data on which they are validated often skew heavily towards a white, male population. Very little attention has been given to validating these models on more diverse populations, in spite of the fact that there is a robust body of academic literature that captures a variety of gendered and racialized pathways to crime \cite{Belknap14,ChesneyLind89,Simpson89}. 
This has significant implications for how we understand the predictive and diagnostic potency of risk assessment tools for diverse populations \cite{ReisigHM06}. For example, scholars have argued that covariates currently included in many regression models for risk prediction are limited to a narrow set of androcentric theories of criminogenic need, to the exclusion of risk and interventional covariates specific to female defendants. The decision of which covariates to add into the regression model is largely driven by how accurate the model's predictions are, not by well-posed questions around how causal the covariates are, potentially excluding covariates that are core drivers of criminal behavior for different subgroups in the population.  For example, some analyses of the LSI-R risk assessment tool have shown that the covariates used for risk prediction are largely informed by a male offender worldview, ignoring well-established ``pathways to crime'' for women (i.e. childhood sexual abuse, substance addiction, and lack of employment) as a crucial interventional class of covariates specific to women \cite{ReisigHM06}.

\emph{Second}, the set of factors considered in RNR assessments are typically constrained to a narrow set of variables, ones which are amenable to individualized treatment plans and are informed by particular psychological and normative theories of re-offending \cite{HannahMoffat11}. Very little consideration has been given to broader social or structural drivers of crime. Indeed, influential scholars of some of the most widely used risk assessments in use today have explicitly discounted the relevance of sociological factors, such as ethnicity and socioeconomic status, because those factors are viewed as static and challenging to intervene upon. These researchers claim that it is a myth to think that the "roots of crime are buried deep in structured inequality," arguing that psychological factors, such as antisocial cognitive patterns and beliefs, have demonstrated much stronger statistical significance when it comes to inter-individual variations in re-arrest within the defendant population. \cite{AndrewsB10, prins2017can}.

Yet, as Prins and Reich have pointed out, these scholars often conflate "causes of crime" with "causes of individual differences in crime." \cite{prins2017can}. Such heavy emphasis on psycho-social factors may limit our ability to understand the underlying drivers of criminal activity at the population level, or ask harder questions about why we see persistent disparities in crime and arrest rates across certain populations and geographies \cite{prins2017can}. As Hannah-Moffat has argued, in RNR assessments an offender's needs are stripped of their broader social context, and are framed largely in terms of poor choices or moral and psychological deficiencies that can be treated through ``re-education'' at the individual level \cite{HannahMoffat11}. This approach tends to obfuscate the significance of underlying factors that are highly prevalent for the population of interest. As Prins and Reich argue, if such structural factors are, "nearly ubiquitous for individuals who become involved in the criminal justice system, we would not expect them to be predictive of inter-individual variation in arrest" \cite{prins2017can}.

The narrow theoretical focus of risk assessments stems in part from the fact that regression is ill-suited to test and differentiate competing models of criminal behavior. This has led to a fairly narrow conceptualization of criminogenic need that is limited to attributes which demonstrate a statistically significant relationship to future recidivism rates in a regression model. Other common sense needs factors for which a statistically significant relationship could not be established, such as mental health, have been excluded from these models.

Within a regression framework, one might address this issue by identifying competing theories of criminogenic behavior and running two different regression models to see which one is more predictive. But if two different regression models were estimated using different guiding criminogenic theories for different subgroups in the population, one has to contend with the real possibility of the models arriving at different risk predictions, leaving no principled way to choose the prediction of one model over another.

\emph{Finally}, it is challenging to differentiate intermediate outcomes from covariates in a regression based model sans a causal inference framework \cite{Greiner08}. This is important because without establishing a causal relationship between a given covariate to the risk of recidivism, we run the risk of misinterpreting an intermediate outcome as a causal driver of crime. This is deeply significant if we are to use risk assessments as the basis for identifying effective interventions to address the underlying drivers of crime. Associative relationships between covariates used in a regression model and the outcome variable of interest need not be causal; such covariates might be misconstrued as an intermediate outcome that is causal of criminal behavior. Prins and Reich underline this important point by means of a compelling example using a directed acyclic graph. They show how intensive policing causes more re-arrests and anti law-enforcement resentment (``antisocial cognition'') at the same time. While antisocial cognition might strongly predict re-arrest, it is not typically a causal driver of re-arrest. Intervening to tackle antisocial cognition without addressing the root cause of excessive and intensive policing will not reduce re-arrests under such a model - antisocial cognition is itself caused by increased intensive policing and must not be misinterpreted as a causal driver of crime. 

It is important to distinguish machine learning from causal inference. Machine learning leverages concepts of pattern recognition, prediction, abstraction and generalization, while a causal inference framework is based on estimating the effect of an intervention on a outcome variable of interest. Machine learning algorithms leverage the same principles as regression for predicting an outcome variable, but do so by expanding the set of covariates for prediction. In machine learning, one wants to obtain a generalized model that appropriately fits the data at hand. This is achieved in machine learning using a variety of advanced statistical techniques such as kernel transformations and parameter sweeping, which enable complex interactions between covariates to be captured (but not necessarily understood) for the purposes of increased predictive accuracy. In a sense, machine learning treats prediction as sacrosanct -- it is not important why a given set of covariates are predictive, so long as they are.

%
%

Machine learning algorithms used for prediction of risk scores should be seen as incremental extensions of regression-based methods, ones that will amplify, rather than mitigate, the challenges we describe above. Though the vast majority of risk prediction models are based on regression, there has been an increased interest in using contemporary ``supervised'' machine learning methods in this setting. In supervised machine learning, a given set of covariates, also referred to as features, are used in a model to predict the outcome variable of interest. An analyst estimating a model to predict risk scores has many types of algorithms to choose from. Many of these algorithms use advanced statistical methods to transform the space of input features into a higher order space that is often difficult to interpret, even by the analyst estimating the model.

In a supervised machine learning driven prediction setting, the principal focus is on achieving reliable and accurate prediction. The question of what constitutes the input space of features is secondary to achieving the best predictions possible. Often a machine learning engineer will simply add or remove individual features, reestimate a new model with the changed feature spaces and check for the predictive performance of the model \cite{james2013an}. There is often no other rationale given for including and excluding individual features. Similar to regression, the set of input features are merely correlational and not causal of the outcome variable. Recent work on interpretable supervised learning algorithms for risk score prediction in the criminal justice system attempts to explain the logic used by the model in computing the risk for a new defendant, but this should be seen as incremental to current practices, as it fails to answer the question of which features are causal drivers of criminal behavior.

Furthermore, such a supervised setting, even with better model interpretability \cite{ZengUR17}, is not a principled way of determining how much a given variable needs to change to reduce the risk of future crime. For example, determining number of hours of cognitive behavioral therapy given to incarcerated individuals to address their criminogenic psychopathology (a common intervenable variable) is difficult to compute using these methods devoid a causal inference framework. In short, without a causal inference framework, contemporary supervised learning algorithms used for risk prediction are not useful for targeted intervention. The increased attention and excitement around machine learning and artificial intelligence, coupled with the advanced statistical techniques they make use of has given them the veneer of appearing ``scientific,'' accompanied by descriptive language that often subtly connotes causality, when in reality these models do not even broach causality.

%% file: towards-causal.tex
\section{Towards a Causal Framework for Risk Assessment}
\label{sec:towards-causal}
In contrast to regression and machine learning, statistical causal inference \cite{imbens2015causal} is a framework that is used to establish causal relationships between covariates and the outcome variable of interest. This is achieved through the design of experimental conditions in which covariates are altered systematically to see if the alteration produces effect changes in the outcome variable. The key difference between causal inference and regression-like methods is the ability for one to change values of a covariate for the purpose of examining whether it causes the outcome variable. Causal inference is best suited for the design of interventions that reduce the risk of future crime.

Yet regression, and to a limited extent predictive machine learning, dominate the landscape of risk assessments used for both prediction and intervention. This gives rise to a series of challenges that limit the justice's systems ability to effectively intervene to lower the risk of recidivism.  A causal inference experiment produces new data to be analyzed for the purpose of establishing causality, but in regression and machine learning, one is limited to historical data that has already been captured, potentially excluding important variables crucial for effective intervention.

Unlike regression and predictive supervised machine learning models, a causal inference framework is better suited to answer questions of ``what interventions work'' in the criminal justice system, particularly those aimed at decreasing well-specified risks, such as recidivism and failure to appear in court. For example consider the potential outcomes causal inference framework. In this approach causality is inferred by randomly assigning individuals or groups, referred to as units, to an intervention or treatment. Each unit subjected to a treatment may realize an outcome of interest, and upon receiving no treatment may realize an alternate outcome, also known as the counter-factual. Randomly assigning units to both the treatment and the control and comparing the potential outcome after the application of treatment and control gives a measure of the causal effect of the chosen intervention.

When performed under an experimental setting, this is known as a randomized control trial (RCT) and is considered the gold standard for measuring the causal effects of a given treatment, under a set of conditions for a defined unit \cite{imbens2015causal}. For example, a randomized-controlled trial in Philadelphia found that community-based low intensity supervision program of low-risk offenders was more effective than high-intensity law-enforcement supervision in reducing new criminal activity of the offenders \cite{barnes2010low}.

In the criminal justice system it may not always be possible to practically obtain a potential outcome for both the treatment and the control of the same unit. Beyond technical challenges with the data, researchers may encounter resistance from the legal profession, which has been reticent to adopt rigorous empirical methods to infer the effectiveness of interventions in the criminal justice system. Greiner and Matthews \cite{greiner2016randomized} point to the fact that the first RCT's in the medical and legal fields were conducted around the same time, in the early 1930s. Yet today, the legal field overwhelmingly lags medical research in the use of scientific causal methods to evaluate the effectiveness of interventions.  

Causal inference in cases where it is not possible to do an RCT is usually derived from applying the same concepts to observational data, with donor pooling and other matching methods to fill the counter-factual values for each unit based on identifying similar units receiving the two different interventions \cite{rubin2006matched}. For example, a recent observational study evaluated the effectiveness of different forms of electronic monitoring (EM) technologies for released defendants in the state of Florida. Using propensity-score matching to group unit (defendant) covariates, the study found that while EM reduced new criminal activity (NCA) across all demographics, the decrease was the lowest for violent offenders \cite{bales2010quantitative}. In addition, electronic monitoring was found to reduce NCA more than the use of global-positioning system (GPS) monitoring. 

The potential outcomes model offers several structural advantages over regression-based methods for evaluating which interventions work in the criminal justice system \cite{Greiner08}. \emph{First}, one is able to measure the impact of timing and duration of the applied intervention in a causal inference framework, an advantage severely lacking in regression-based methods. For example, given the limited resources for behavioral therapy in correctional settings, research has shown that the timing of the initiation of behavioral therapy has an effect not only on prison conduct of defendants, but also on the risk of recidivism.  In the potential outcomes framework, a randomized experiment can test when it might be a good time to initiate behavioral therapy as an intervention \cite{Duwe2017}.

\emph{Second}, random assignment of units to treatments ensures a ``balance'' of the covariates, thereby isolating the applied treatment as the causal driver (or not). \emph{Third}, causal inference insists that the analyst carefully separates covariates that are not impacted by the treatment from intermediate outcomes that are indeed impacted by the treatment. This has serious implications for using risk assessments as a diagnostic tool, especially for high-risk groups whose risk stems from structural inequality.

This doesn't mean that there is no room for machine learning in the development of diagnostic risk assessments. Unsupervised and semi-supervised machine learning algorithms are well suited for surfacing important causal questions regarding highly correlational covariates. Such causal inference questions are much better addressed by either performing inference on observational data, or even better, the design of a randomized experiment when possible. For example, Kim and Duwe show \cite{DuweKim16} how supervised machine learning models outperform classical regression models for predicting recidivism. Yet, these models are difficult to interpret, even though they yield more accurate predictions. Rather than using machine learning for prediction, these methods could be used to identify features that are highly predictive of recidivism, in order to inform hypotheses on interventions (and their timing) that can then be tested using causal inference.

%% file: conclusion.tex
\section{Implications and Conclusion}
\label{sec:conclusion}
In recent times, predictive risk assessments have garnered unprecedented interest, as individuals from across the political spectrum seek to achieve a diverse set of criminal justice reforms with the help of data-driven actuarial tools. The promise of these tools is frequently characterized in terms of the affordances they offer with regard to cutting-edge technological advances in predictive data analysis, in fields such as machine learning. Risk assessments have been sold as an innovative means of achieving efficiency, neutrality, and fairness in a system that has been plagued by the implicit bias of individual decision makers (i.e. judges) and flawed organizational risk mitigation practices (i.e. money-bail) \cite{pji}.

However, the reality is that most risk assessments in use today are built using statistical methods that are based on regression. Little has changed since the 1970's when the first statistically validated actuarial tools were built using regression algorithms. Even in instances where new methods commonly associated with ``artificial intelligence'' are being developed for risk assessments, these innovations are really incremental extensions of prior statistical methods used for forecasting future criminal events.

Risk assessment has a long and rich history in the criminal justice system. Yet, in current debates surrounding the ethical use of these tools, this larger historical context is often lost. Fairness debates surrounding risk assessment end up in conceptual cul-de-sacs, whereby trade-offs regarding things like ``accuracy equity'' and ``predictive parity'' are weighed in lieu of asking deeper questions about what the purpose of these tools is and should be in the first place \cite{FloresBL16,KleinbergMR17,Gong16}. Those deeper questions include asking how actuarial tools might support or undermine crime prevention interventions.

We argue that when risk assessments are used primarily as a predictive technology, they fuel harmful trends towards mass incarceration and growing inequality in the justice system. Predictive risk assessments offer little guidance on how to effectively intervene to lower risk. When predictive accuracy is the primary metric along which these technologies are evaluated, the system misses opportunities to explore a deeper set of questions surrounding the way its administrators can use data as part of a reflexive practice of testing hypotheses in the service of achieving near and long term goals.

We argue for a shift away from predictive technologies, towards diagnostic methods that will help us to understand the criminogenic effects of the criminal justice system itself, as well as evaluate the effectiveness of interventions designed to interrupt cycles of crime. In contrast to the current emphasis on machine learning techniques that offer no grounded way of understanding the underlying drivers of crime, these methods should be based in a more rigorous approach that incorporates both qualitative and quantitative data analysis.

To this end, we argue that risk assessments should be conceived of as a diagnostic tool that can be used to understand the underlying social, economic and psychological drivers of crime. We posit that causal inference offers the best framework for pursuing these goals. More work should be done to examine how these quantitative methods might be supplemented by more qualitative practices of knowledge production. For example, Paluck argues that ethnographic methods can enable researchers to move beyond average treatment effects, to more deeply understand the underlying mechanisms of a causal effect via causal inference \cite{Paluck10}. Moreover, Elish and boyd have pointed out that quantitative researchers would benefit immensely from the rich frameworks for reflexivity found in more qualitative data practices, such as ethnography \cite{ElishB17}. This is of paramount importance for data applications in the criminal justice system, where researchers should constantly re-examine the way their research practices might  influence and warp the outcomes of their work.

Though causal inference through randomized experiments has been touted as the gold standard for evaluating criminogenic interventions, their practical application is much more sparse \cite{weisburd2000randomized, greiner2016randomized}. A large scale survey of intervention evaluations in the US criminal justice system found that only $16\%$ of all interventions were evaluated using randomized experiments \cite{sherman2002}. Randomized experiments to infer causal relationships between interventions and recidivism are seen as difficult to implement in practice, given resource constraints in correctional facilities. Others have raised ethical challenges around subjecting a vulnerable population to a control arm of a randomized experiment \cite{weisburd2000randomized}. These concerns and challenges mirror challenges in medical research, where the testing of medical devices, drugs and therapies can often have life and death implications. Yet, there are well-established and mature frameworks on ethical experimentation and causal inference in clinical medicine. It is inconceivable today to imagine the introduction of a new drug without RCTs -- the Food and Drug Administration mandates a framework for furnishing evidence of a new drug's effectiveness in four phases. Guidelines and rules for ethical experimentation on patients from funding bodies such as the National Institutes of Health, combined with Institutional Review Board (IRB) oversight at host institutions are the norm for causal inference in medical research. Scholars such as Greiner \cite{greiner2016randomized} have cast the lack of causal inference in the law as a problem of political will. 
%
%
%
%

In the criminal justice system, many new interventions, such as the timing, type and doses of psychological therapy both within and outside of correctional facilities can be tested using randomized experiments. In cases where a randomized experiment is either not feasible or is ethically fraught with serious potential risks, deriving causal inference from observational data should be considered \cite{Greiner08}. There is a growing body of work that deploys observational methods in the field of criminal justice. For instance, Dobbie, Goldin and Yang show via quasi-experimental observational study design that pretrial detention weakens defendant's bargaining position in plea negotiations and a conviction also reduces their future participation in the the formal labor market \cite{dobbie2016effects}. Just in the last year, four other observational causal inference studies exploring the downstream consequences of pretrial detention were released \cite{gupta2016heavy,heaton2017downstream,leslie2016unintended, stevenson2016distortion}.

In addition, Lum and Yang \cite{lum2005evaluation} have highlighted structural challenges that also hamper the use of causal inference studies, such as the lack of early mentorship available for empirical criminology analysts interested in using these methods, as well as a shift of research funding away from randomized experiments to quick statistical analysis. The relative ease with which most regression and supervised prediction algorithms can be estimated makes them all the more convenient to use. It's no surprise that machine learning has gained momentum over causal inference, at a time when convenience is combined with the growing hype around the use of machine learning to solve some of our most intractable social problems. Nevertheless, careful design of a causal inference setup, both experimental and observational, offer significant benefits for the process of discovering and validating criminogenic interventions.

Data-driven tools provide an immense opportunity for us to pursue goals of fair punishment and future crime prevention. But this requires us to move away from merely tacking on intervenable variables to risk covariates for predictive models, and towards the use of empirically-grounded tools to help understand and respond to the underlying drivers of crime, both individually and systemically.


%% file: acks.tex
The research leading to these results has received funding from the Ethics and Governance of Artificial Intelligence Fund.

%% file: fatml-paper.bbl

\begin{thebibliography}{74}


\ifx \showCODEN    \undefined \def \showCODEN     #1{\unskip}     \fi
\ifx \showDOI      \undefined \def \showDOI       #1{#1}\fi
\ifx \showISBNx    \undefined \def \showISBNx     #1{\unskip}     \fi
\ifx \showISBNxiii \undefined \def \showISBNxiii  #1{\unskip}     \fi
\ifx \showISSN     \undefined \def \showISSN      #1{\unskip}     \fi
\ifx \showLCCN     \undefined \def \showLCCN      #1{\unskip}     \fi
\ifx \shownote     \undefined \def \shownote      #1{#1}          \fi
\ifx \showarticletitle \undefined \def \showarticletitle #1{#1}   \fi
\ifx \showURL      \undefined \def \showURL       {\relax}        \fi
\providecommand\bibfield[2]{#2}
\providecommand\bibinfo[2]{#2}
\providecommand\natexlab[1]{#1}
\providecommand\showeprint[2][]{arXiv:#2}

\bibitem[\protect\citeauthoryear{Andrews and Bonta}{Andrews and Bonta}{2010}]%
        {AndrewsB10}
\bibfield{author}{\bibinfo{person}{D.A. Andrews} {and} \bibinfo{person}{J.
  Bonta}.} \bibinfo{year}{2010}\natexlab{}.
\newblock \bibinfo{booktitle}{{\em The Psychology of Criminal Conduct}}.
\newblock \bibinfo{publisher}{Lexis Nexis/Anderson Pub.}
\newblock
\showISBNx{9781422463291}
\showLCCN{2010000283}


\bibitem[\protect\citeauthoryear{Andrews, Bonta, and Hoge}{Andrews
  et~al\mbox{.}}{1990}]%
        {AndrewsBH90}
\bibfield{author}{\bibinfo{person}{D.~A. Andrews}, \bibinfo{person}{James
  Bonta}, {and} \bibinfo{person}{R.~D. Hoge}.} \bibinfo{year}{1990}\natexlab{}.
\newblock \showarticletitle{Classification for Effective Rehabilitation}.
\newblock \bibinfo{journal}{{\em Criminal Justice and Behavior\/}}
  \bibinfo{volume}{17}, \bibinfo{number}{1} (\bibinfo{year}{1990}),
  \bibinfo{pages}{19--52}.
\newblock
\showDOI{%
\url{https://doi.org/10.1177/0093854890017001004}}


\bibitem[\protect\citeauthoryear{Andrews, Bonta, and Wormith}{Andrews
  et~al\mbox{.}}{2006}]%
        {AndrewsBW06}
\bibfield{author}{\bibinfo{person}{D.~A. Andrews}, \bibinfo{person}{James
  Bonta}, {and} \bibinfo{person}{J.~Stephen Wormith}.}
  \bibinfo{year}{2006}\natexlab{}.
\newblock \showarticletitle{The Recent Past and Near Future of Risk and/or Need
  Assessment}.
\newblock \bibinfo{journal}{{\em Crime \& Delinquency\/}} \bibinfo{volume}{52},
  \bibinfo{number}{1} (\bibinfo{year}{2006}), \bibinfo{pages}{7--27}.
\newblock
\showDOI{%
\url{https://doi.org/10.1177/0011128705281756}}


\bibitem[\protect\citeauthoryear{Angwin, Larson, Mattu, and Kirchner}{Angwin
  et~al\mbox{.}}{2016}]%
        {AngwinLMK16}
\bibfield{author}{\bibinfo{person}{Julia Angwin}, \bibinfo{person}{Jeff
  Larson}, \bibinfo{person}{Surya Mattu}, {and} \bibinfo{person}{Lauren
  Kirchner}.} \bibinfo{year}{2016}\natexlab{}.
\newblock \showarticletitle{Machine Bias: There's software used across the
  country to predict future criminals. And it's biased against blacks.}
\newblock \bibinfo{journal}{{\em ProPublica\/}} (\bibinfo{year}{2016}).
\newblock


\bibitem[\protect\citeauthoryear{Bales, Mann, Blomberg, Gaes, Barrick,
  Dhungana, and McManus}{Bales et~al\mbox{.}}{2010}]%
        {bales2010quantitative}
\bibfield{author}{\bibinfo{person}{William Bales}, \bibinfo{person}{Karen
  Mann}, \bibinfo{person}{Thomas Blomberg}, \bibinfo{person}{Gerry Gaes},
  \bibinfo{person}{Kelle Barrick}, \bibinfo{person}{Karla Dhungana}, {and}
  \bibinfo{person}{Brian McManus}.} \bibinfo{year}{2010}\natexlab{}.
\newblock \showarticletitle{Quantitative and Qualitative Assessment of
  Electronic Monitoring}.
\newblock  (\bibinfo{year}{2010}).
\newblock


\bibitem[\protect\citeauthoryear{Barnes, Ahlman, Gill, Sherman, Kurtz, and
  Malvestuto}{Barnes et~al\mbox{.}}{2010}]%
        {barnes2010low}
\bibfield{author}{\bibinfo{person}{Geoffrey~C Barnes}, \bibinfo{person}{Lindsay
  Ahlman}, \bibinfo{person}{Charlotte Gill}, \bibinfo{person}{Lawrence~W
  Sherman}, \bibinfo{person}{Ellen Kurtz}, {and} \bibinfo{person}{Robert
  Malvestuto}.} \bibinfo{year}{2010}\natexlab{}.
\newblock \showarticletitle{Low-intensity community supervision for low-risk
  offenders: a randomized, controlled trial}.
\newblock \bibinfo{journal}{{\em Journal of Experimental Criminology\/}}
  \bibinfo{volume}{6}, \bibinfo{number}{2} (\bibinfo{year}{2010}),
  \bibinfo{pages}{159--189}.
\newblock


\bibitem[\protect\citeauthoryear{Barocas and Selbst}{Barocas and
  Selbst}{2016}]%
        {BarocasS14}
\bibfield{author}{\bibinfo{person}{Solon Barocas} {and}
  \bibinfo{person}{Andrew~D. Selbst}.} \bibinfo{year}{2016}\natexlab{}.
\newblock \showarticletitle{Big Data's Disparate Impact}.
\newblock \bibinfo{journal}{{\em California Law Review\/}}
  \bibinfo{volume}{104} (\bibinfo{year}{2016}).
\newblock
Issue 3.
\showDOI{%
\url{https://doi.org/10.15779/Z38BG31}}


\bibitem[\protect\citeauthoryear{Beck}{Beck}{1992}]%
        {Beck92}
\bibfield{author}{\bibinfo{person}{Ulrich Beck}.}
  \bibinfo{year}{1992}\natexlab{}.
\newblock \bibinfo{booktitle}{{\em Risk Society: Towards a New Modernity}}.
\newblock \bibinfo{publisher}{SAGE Publications Ltd}. 272 pages.
\newblock


\bibitem[\protect\citeauthoryear{Belknap}{Belknap}{2014}]%
        {Belknap14}
\bibfield{author}{\bibinfo{person}{J. Belknap}.}
  \bibinfo{year}{2014}\natexlab{}.
\newblock \bibinfo{booktitle}{{\em The Invisible Woman: Gender, Crime, and
  Justice}}.
\newblock \bibinfo{publisher}{Cengage Learning}.
\newblock
\showISBNx{9781305175730}


\bibitem[\protect\citeauthoryear{Berk, Heidari, Jabbari, Kearns, and Roth}{Berk
  et~al\mbox{.}}{2017}]%
        {BerkHJKR17}
\bibfield{author}{\bibinfo{person}{Richard Berk}, \bibinfo{person}{Hoda
  Heidari}, \bibinfo{person}{Shahin Jabbari}, \bibinfo{person}{Michael Kearns},
  {and} \bibinfo{person}{Aaron Roth}.} \bibinfo{year}{2017}\natexlab{}.
\newblock \bibinfo{booktitle}{{\em Fairness in Criminal Justice Risk
  Assessments: The State of the Art}}.
\newblock \bibinfo{type}{Working Paper} 2017-1.0.
  \bibinfo{institution}{University of Pennsylvania, Department of Criminology}.
\newblock
\showURL{%
\url{https://arxiv.org/abs/1703.09207}}


\bibitem[\protect\citeauthoryear{Biggs}{Biggs}{2016}]%
        {Biggs16}
\bibfield{author}{\bibinfo{person}{John Biggs}.}
  \bibinfo{year}{2016}\natexlab{}.
\newblock \bibinfo{title}{Naborly lets landlords screen tenants automagically}.
\newblock   (\bibinfo{year}{2016}).
\newblock
\showURL{%
\url{https://techcrunch.com/2016/08/15/naborly-lets-landlords-screen-tenants-automagically/}}


\bibitem[\protect\citeauthoryear{Chesney-Lind}{Chesney-Lind}{1989}]%
        {ChesneyLind89}
\bibfield{author}{\bibinfo{person}{Meda Chesney-Lind}.}
  \bibinfo{year}{1989}\natexlab{}.
\newblock \showarticletitle{Girls' Crime and Woman's Place: Toward a Feminist
  Model of Female Delinquency}.
\newblock \bibinfo{journal}{{\em Crime \& Delinquency\/}} \bibinfo{volume}{35},
  \bibinfo{number}{1} (\bibinfo{year}{1989}), \bibinfo{pages}{5--29}.
\newblock
\showDOI{%
\url{https://doi.org/10.1177/0011128789035001002}}


\bibitem[\protect\citeauthoryear{Citron}{Citron}{2016}]%
        {Citron16}
\bibfield{author}{\bibinfo{person}{Danielle Citron}.}
  \bibinfo{year}{2016}\natexlab{}.
\newblock \bibinfo{title}{(Un)Fairness of Risk Scores in Criminal Sentencing}.
\newblock   (\bibinfo{year}{2016}).
\newblock
\showURL{%
\url{https://www.forbes.com/sites/daniellecitron/2016/07/13/unfairness-of-risk-scores-in-criminal-sentencing/}}


\bibitem[\protect\citeauthoryear{Cullen and Genderau}{Cullen and
  Genderau}{2001}]%
        {CullenG01}
\bibfield{author}{\bibinfo{person}{Francis~T. Cullen} {and}
  \bibinfo{person}{Paul Genderau}.} \bibinfo{year}{2001}\natexlab{}.
\newblock \showarticletitle{From Nothing Works to What Works: Changing
  Professional Ideology in the 21st Century}.
\newblock \bibinfo{journal}{{\em The Prison Journal\/}} \bibinfo{volume}{81},
  \bibinfo{number}{3} (\bibinfo{year}{2001}), \bibinfo{pages}{313--338}.
\newblock
\showDOI{%
\url{https://doi.org/10.1177/0032885501081003002}}


\bibitem[\protect\citeauthoryear{Cullen, Jonson, and Nagin}{Cullen
  et~al\mbox{.}}{2011}]%
        {CullenJN11}
\bibfield{author}{\bibinfo{person}{Francis~T. Cullen},
  \bibinfo{person}{Cheryl~Lero Jonson}, {and} \bibinfo{person}{Daniel~S.
  Nagin}.} \bibinfo{year}{2011}\natexlab{}.
\newblock \showarticletitle{Prisons Do Not Reduce Recidivism}.
\newblock \bibinfo{journal}{{\em The Prison Journal\/}} \bibinfo{volume}{91},
  \bibinfo{number}{3} (\bibinfo{year}{2011}), \bibinfo{pages}{48S--65S}.
\newblock
\showDOI{%
\url{https://doi.org/10.1177/0032885511415224}}


\bibitem[\protect\citeauthoryear{DeFina and Hannon}{DeFina and Hannon}{2010}]%
        {DeFinaH10}
\bibfield{author}{\bibinfo{person}{Robert DeFina} {and} \bibinfo{person}{Lance
  Hannon}.} \bibinfo{year}{2010}\natexlab{}.
\newblock \showarticletitle{For incapacitation, there is no time like the
  present: The lagged effects of prisoner reentry on property and violent crime
  rates}.
\newblock \bibinfo{journal}{{\em Social Science Research\/}}
  \bibinfo{volume}{39}, \bibinfo{number}{6} (\bibinfo{year}{2010}),
  \bibinfo{pages}{1004--1014}.
\newblock
\showDOI{%
\url{https://doi.org/10.1016/j.ssresearch.2010.08.001}}


\bibitem[\protect\citeauthoryear{Desmarais and Singh}{Desmarais and
  Singh}{2013}]%
        {DesmaraisS13}
\bibfield{author}{\bibinfo{person}{Sarah~L. Desmarais} {and}
  \bibinfo{person}{Jay~P. Singh}.} \bibinfo{year}{2013}\natexlab{}.
\newblock \bibinfo{booktitle}{{\em Risk Assessment Instruments Validated and
  Implemented in Correctional Settings in the United States}}.
\newblock \bibinfo{type}{{T}echnical {R}eport}. \bibinfo{institution}{The
  Council of State Governments Justice Center}.
\newblock


\bibitem[\protect\citeauthoryear{Dobbie, Goldin, and Yang}{Dobbie
  et~al\mbox{.}}{2016}]%
        {dobbie2016effects}
\bibfield{author}{\bibinfo{person}{Will Dobbie}, \bibinfo{person}{Jacob
  Goldin}, {and} \bibinfo{person}{Crystal Yang}.}
  \bibinfo{year}{2016}\natexlab{}.
\newblock \bibinfo{booktitle}{{\em The effects of pre-trial detention on
  conviction, future crime, and employment: Evidence from randomly assigned
  judges}}.
\newblock \bibinfo{type}{{T}echnical {R}eport}. \bibinfo{institution}{National
  Bureau of Economic Research}.
\newblock


\bibitem[\protect\citeauthoryear{Duwe}{Duwe}{2017}]%
        {Duwe2017}
\bibfield{author}{\bibinfo{person}{Grant Duwe}.}
  \bibinfo{year}{2017}\natexlab{}.
\newblock \bibinfo{title}{Timing and Sequence of Correctional Programming}.
\newblock   (\bibinfo{year}{2017}).
\newblock
\showURL{%
\url{http://www.crj.org/assets/2017/09/Timing-and-Sequencing-of-Correctional-Programming-Duwe.pdf}}


\bibitem[\protect\citeauthoryear{Duwe and Kim}{Duwe and Kim}{2016}]%
        {DuweKim16}
\bibfield{author}{\bibinfo{person}{Grant Duwe} {and} \bibinfo{person}{KiDeuk
  Kim}.} \bibinfo{year}{2016}\natexlab{}.
\newblock \showarticletitle{Sacrificing Accuracy for Transparency in Recidivism
  Risk Assessment: The Impact of Classification Method on Predictive
  Performance}.
\newblock \bibinfo{journal}{{\em Corrections\/}} \bibinfo{volume}{1},
  \bibinfo{number}{3} (\bibinfo{year}{2016}), \bibinfo{pages}{155--176}.
\newblock
\showDOI{%
\url{https://doi.org/10.1080/23774657.2016.1178083}}


\bibitem[\protect\citeauthoryear{Economist}{Economist}{2017}]%
        {Economist17}
\bibfield{author}{\bibinfo{person}{The Economist}.}
  \bibinfo{year}{2017}\natexlab{}.
\newblock \bibinfo{title}{Machine-learning promises to shake up large swathes
  of finance}.
\newblock   (\bibinfo{year}{2017}).
\newblock
\showURL{%
\url{https://www.economist.com/news/finance-and-economics/21722685-fields-trading-credit-assessment-fraud-prevention-machine-learning}}


\bibitem[\protect\citeauthoryear{Elish and danah boyd}{Elish and danah
  boyd}{2017}]%
        {ElishB17}
\bibfield{author}{\bibinfo{person}{M.~C. Elish} {and} \bibinfo{person}{danah
  boyd}.} \bibinfo{year}{2017}\natexlab{}.
\newblock \showarticletitle{Situating methods in the magic of Big Data and AI}.
\newblock \bibinfo{journal}{{\em Communication Monographs\/}}
  \bibinfo{volume}{0}, \bibinfo{number}{0} (\bibinfo{year}{2017}),
  \bibinfo{pages}{1--24}.
\newblock
\showDOI{%
\url{https://doi.org/10.1080/03637751.2017.1375130}}


\bibitem[\protect\citeauthoryear{Etienne}{Etienne}{2009}]%
        {Etienne09}
\bibfield{author}{\bibinfo{person}{Margareth Etienne}.}
  \bibinfo{year}{2009}\natexlab{}.
\newblock \showarticletitle{Legal and practical implications of evidence-based
  sentencing by judges}.
\newblock \bibinfo{journal}{{\em Chapman Journal of Criminal Justice\/}}
  \bibinfo{volume}{1} (\bibinfo{year}{2009}), \bibinfo{pages}{43--60}.
\newblock


\bibitem[\protect\citeauthoryear{Feeley and Simon}{Feeley and Simon}{1992}]%
        {FeeleyS92}
\bibfield{author}{\bibinfo{person}{Malcolm~M. Feeley} {and}
  \bibinfo{person}{Jonathan Simon}.} \bibinfo{year}{1992}\natexlab{}.
\newblock \showarticletitle{The New Penology: Notes on the Emerging Strategy of
  Corrections and Its Implications}.
\newblock \bibinfo{journal}{{\em Criminology\/}}  \bibinfo{volume}{30}
  (\bibinfo{year}{1992}), \bibinfo{pages}{449--475}.
\newblock
\showURL{%
\url{http://scholarship.law.berkeley.edu/facpubs/718/}}


\bibitem[\protect\citeauthoryear{Fennell and Hall}{Fennell and Hall}{1980}]%
        {FennelH80}
\bibfield{author}{\bibinfo{person}{Stephen~A. Fennell} {and}
  \bibinfo{person}{William~N. Hall}.} \bibinfo{year}{1980}\natexlab{}.
\newblock \showarticletitle{Due Process at Sentencing: An Empirical and Legal
  Analysis of the Disclosure of Presentence Reports in Federal Courts}.
\newblock \bibinfo{journal}{{\em Harvard Law Review\/}} \bibinfo{volume}{93},
  \bibinfo{number}{8} (\bibinfo{year}{1980}), \bibinfo{pages}{1613--1697}.
\newblock
\showISSN{0017811X}


\bibitem[\protect\citeauthoryear{Flores, Bechtel, and Lowenkamp}{Flores
  et~al\mbox{.}}{2016}]%
        {FloresBL16}
\bibfield{author}{\bibinfo{person}{Anthony~W. Flores}, \bibinfo{person}{Kristin
  Bechtel}, {and} \bibinfo{person}{Christopher~T. Lowenkamp}.}
  \bibinfo{year}{2016}\natexlab{}.
\newblock \showarticletitle{False Positives, False Negatives, and False
  Analyses: A Rejoinder to ``Machine Bias: There's Software Used Across the
  Country to Predict Future Criminals. And It's Biased Against Blacks.''}.
\newblock \bibinfo{journal}{{\em Federal Probation Journal\/}}
  (\bibinfo{date}{September} \bibinfo{year}{2016}), \bibinfo{pages}{38--46}.
\newblock
\showURL{%
\url{http://www.uscourts.gov/statistics-reports/publications/federal-probation-journal/federal-probation-journal-september-2016}}


\bibitem[\protect\citeauthoryear{Foundation}{Foundation}{2016}]%
        {arnoldF}
\bibfield{author}{\bibinfo{person}{Arnold Foundation}.}
  \bibinfo{year}{2016}\natexlab{}.
\newblock \showarticletitle{Results from the First Six Months of the Public
  Safety Assessment–Court in Kentucky}.
\newblock \bibinfo{journal}{{\em Report\/}} (\bibinfo{year}{2016}).
\newblock


\bibitem[\protect\citeauthoryear{Gelman}{Gelman}{2007}]%
        {gelman2007data}
\bibfield{author}{\bibinfo{person}{Andrew Gelman}.}
  \bibinfo{year}{2007}\natexlab{}.
\newblock \bibinfo{booktitle}{{\em Data analysis using regression and
  multilevel/hierarchical models}}.
\newblock \bibinfo{publisher}{Cambridge University Press},
  \bibinfo{address}{Cambridge New York}.
\newblock
\showISBNx{978-0521686891}


\bibitem[\protect\citeauthoryear{Gong}{Gong}{2016}]%
        {Gong16}
\bibfield{author}{\bibinfo{person}{Abe Gong}.} \bibinfo{year}{2016}\natexlab{}.
\newblock \bibinfo{title}{Ethics for powerful algorithms (2 of 4)}.
\newblock   (\bibinfo{date}{July} \bibinfo{year}{2016}).
\newblock
\showURL{%
\url{https://medium.com/@AbeGong/ethics-for-powerful-algorithms-2-of-3-5bf750ce4c54}}


\bibitem[\protect\citeauthoryear{Greiner}{Greiner}{2008}]%
        {Greiner08}
\bibfield{author}{\bibinfo{person}{D.~James Greiner}.}
  \bibinfo{year}{2008}\natexlab{}.
\newblock \showarticletitle{Causal Inference In Civil Rights Litigation}.
\newblock \bibinfo{journal}{{\em Harvard Law Review\/}} \bibinfo{volume}{122},
  \bibinfo{number}{2} (\bibinfo{year}{2008}), \bibinfo{pages}{533--598}.
\newblock


\bibitem[\protect\citeauthoryear{Greiner and Matthews}{Greiner and
  Matthews}{2016}]%
        {greiner2016randomized}
\bibfield{author}{\bibinfo{person}{D~James Greiner} {and}
  \bibinfo{person}{Andrea Matthews}.} \bibinfo{year}{2016}\natexlab{}.
\newblock \showarticletitle{Randomized control trials in the United States
  legal profession}.
\newblock \bibinfo{journal}{{\em Annual Review of Law and Social Science\/}}
  \bibinfo{volume}{12} (\bibinfo{year}{2016}), \bibinfo{pages}{295--312}.
\newblock


\bibitem[\protect\citeauthoryear{Gupta, Hansman, and Frenchman}{Gupta
  et~al\mbox{.}}{2016}]%
        {gupta2016heavy}
\bibfield{author}{\bibinfo{person}{Arpit Gupta}, \bibinfo{person}{Christopher
  Hansman}, {and} \bibinfo{person}{Ethan Frenchman}.}
  \bibinfo{year}{2016}\natexlab{}.
\newblock \showarticletitle{The heavy costs of high bail: Evidence from judge
  randomization}.
\newblock \bibinfo{journal}{{\em The Journal of Legal Studies\/}}
  \bibinfo{volume}{45}, \bibinfo{number}{2} (\bibinfo{year}{2016}),
  \bibinfo{pages}{471--505}.
\newblock


\bibitem[\protect\citeauthoryear{Hamilton}{Hamilton}{2015}]%
        {Hamilton15BackToFuture}
\bibfield{author}{\bibinfo{person}{Melissa Hamilton}.}
  \bibinfo{year}{2015}\natexlab{}.
\newblock \showarticletitle{Back to the Future: The Influence of Criminal
  History on Risk Assessment}.
\newblock \bibinfo{journal}{{\em Berkeley Journal of Criminal Law\/}}
  (\bibinfo{year}{2015}).
\newblock
\newblock
\shownote{Forthcoming; SSRN ID: 2555878.}


\bibitem[\protect\citeauthoryear{Hannah-Moffat}{Hannah-Moffat}{2005}]%
        {hannah2005criminogenic}
\bibfield{author}{\bibinfo{person}{Kelly Hannah-Moffat}.}
  \bibinfo{year}{2005}\natexlab{}.
\newblock \showarticletitle{Criminogenic needs and the transformative risk
  subject: Hybridizations of risk/need in penality}.
\newblock \bibinfo{journal}{{\em Punishment \& Society\/}} \bibinfo{volume}{7},
  \bibinfo{number}{1} (\bibinfo{year}{2005}), \bibinfo{pages}{29--51}.
\newblock


\bibitem[\protect\citeauthoryear{Hannah-Moffat}{Hannah-Moffat}{2011}]%
        {HannahMoffat11}
\bibfield{author}{\bibinfo{person}{Kelly Hannah-Moffat}.}
  \bibinfo{year}{2011}\natexlab{}.
\newblock \showarticletitle{Sacrosanct or Flawed: Risk, Accountability and
  Gender-responsive Penal Politics}.
\newblock \bibinfo{journal}{{\em Current Issues in Criminal Justice\/}}
  \bibinfo{volume}{22} (\bibinfo{date}{March} \bibinfo{year}{2011}),
  \bibinfo{pages}{193--215}.
\newblock
Issue 3.


\bibitem[\protect\citeauthoryear{Hannah-Moffat}{Hannah-Moffat}{2013}]%
        {HannahMoffat13}
\bibfield{author}{\bibinfo{person}{Kelly Hannah-Moffat}.}
  \bibinfo{year}{2013}\natexlab{}.
\newblock \showarticletitle{Actuarial Sentencing: An ``Unsettled''
  Proposition}.
\newblock \bibinfo{journal}{{\em Justice Quarterly\/}} \bibinfo{volume}{30},
  \bibinfo{number}{2} (\bibinfo{year}{2013}), \bibinfo{pages}{270--296}.
\newblock
\showDOI{%
\url{https://doi.org/10.1080/07418825.2012.682603}}


\bibitem[\protect\citeauthoryear{Hannah-Moffat}{Hannah-Moffat}{2015}]%
        {HannahMoffat15}
\bibfield{author}{\bibinfo{person}{Kelly Hannah-Moffat}.}
  \bibinfo{year}{2015}\natexlab{}.
\newblock \showarticletitle{The Uncertainties of Risk Assessment: Partiality,
  Transparency, and Just Decisions}.
\newblock \bibinfo{journal}{{\em Federal Sentencing Reporter\/}}
  \bibinfo{volume}{27}, \bibinfo{number}{4} (\bibinfo{date}{April}
  \bibinfo{year}{2015}), \bibinfo{pages}{244--247}.
\newblock
\showDOI{%
\url{https://doi.org/10.1525/fsr.2015.27.4.244}}


\bibitem[\protect\citeauthoryear{Harcourt}{Harcourt}{2010}]%
        {harcourt2010risk}
\bibfield{author}{\bibinfo{person}{Bernard~E Harcourt}.}
  \bibinfo{year}{2010}\natexlab{}.
\newblock \showarticletitle{Risk as a proxy for race}.
\newblock  (\bibinfo{year}{2010}).
\newblock


\bibitem[\protect\citeauthoryear{Harcourt}{Harcourt}{2015}]%
        {Harcourt15}
\bibfield{author}{\bibinfo{person}{Bernard~E. Harcourt}.}
  \bibinfo{year}{2015}\natexlab{}.
\newblock \showarticletitle{Risk as a Proxy for Race: The Dangers of Risk
  Assessment}.
\newblock \bibinfo{journal}{{\em Federal Sentencing Reporter\/}}
  \bibinfo{volume}{27}, \bibinfo{number}{4} (\bibinfo{date}{April}
  \bibinfo{year}{2015}), \bibinfo{pages}{237--243}.
\newblock
\showDOI{%
\url{https://doi.org/10.1525/fsr.2015.27.4.237}}


\bibitem[\protect\citeauthoryear{Hardy}{Hardy}{2014}]%
        {Hardy14}
\bibfield{author}{\bibinfo{person}{Mark Hardy}.}
  \bibinfo{year}{2014}\natexlab{}.
\newblock \showarticletitle{Practitioner perspectives on risk: Using
  governmentality to understand contemporary probation practice}.
\newblock \bibinfo{journal}{{\em European Journal of Criminology\/}}
  \bibinfo{volume}{11}, \bibinfo{number}{3} (\bibinfo{year}{2014}),
  \bibinfo{pages}{303--318}.
\newblock
\showDOI{%
\url{https://doi.org/10.1177/1477370813495758}}


\bibitem[\protect\citeauthoryear{Heaton, Mayson, and Stevenson}{Heaton
  et~al\mbox{.}}{2017}]%
        {heaton2017downstream}
\bibfield{author}{\bibinfo{person}{Paul Heaton}, \bibinfo{person}{Sandra
  Mayson}, {and} \bibinfo{person}{Megan Stevenson}.}
  \bibinfo{year}{2017}\natexlab{}.
\newblock \showarticletitle{The downstream consequences of misdemeanor pretrial
  detention}.
\newblock \bibinfo{journal}{{\em Stan. L. Rev.\/}}  \bibinfo{volume}{69}
  (\bibinfo{year}{2017}), \bibinfo{pages}{711}.
\newblock


\bibitem[\protect\citeauthoryear{Imbens}{Imbens}{2015}]%
        {imbens2015causal}
\bibfield{author}{\bibinfo{person}{Guido Imbens}.}
  \bibinfo{year}{2015}\natexlab{}.
\newblock \bibinfo{booktitle}{{\em Causal inference for statistics, social, and
  biomedical sciences : an introduction}}.
\newblock \bibinfo{publisher}{Cambridge University Press},
  \bibinfo{address}{New York}.
\newblock
\showISBNx{978-0521885881}


\bibitem[\protect\citeauthoryear{Institute}{Institute}{2017}]%
        {pji}
\bibfield{author}{\bibinfo{person}{Pretrial~Justice Institute}.}
  \bibinfo{year}{2017}\natexlab{}.
\newblock \showarticletitle{Pretrial Risk Assessments Can Produce Race-Neutral
  Results}.
\newblock \bibinfo{journal}{{\em Report\/}} (\bibinfo{year}{2017}).
\newblock


\bibitem[\protect\citeauthoryear{James}{James}{2013}]%
        {james2013an}
\bibfield{author}{\bibinfo{person}{Gareth James}.}
  \bibinfo{year}{2013}\natexlab{}.
\newblock \bibinfo{booktitle}{{\em An introduction to statistical learning :
  with applications in R}}.
\newblock \bibinfo{publisher}{Springer}, \bibinfo{address}{New York, NY}.
\newblock
\showISBNx{978-1461471370}


\bibitem[\protect\citeauthoryear{Kitchin}{Kitchin}{2014}]%
        {Kitchin14}
\bibfield{author}{\bibinfo{person}{Rob Kitchin}.}
  \bibinfo{year}{2014}\natexlab{}.
\newblock \showarticletitle{Big Data, new epistemologies and paradigm shifts}.
\newblock \bibinfo{journal}{{\em Big Data \& Society\/}} \bibinfo{volume}{1},
  \bibinfo{number}{1} (\bibinfo{year}{2014}),
  \bibinfo{pages}{2053951714528481}.
\newblock
\showDOI{%
\url{https://doi.org/10.1177/2053951714528481}}


\bibitem[\protect\citeauthoryear{Kleinberg, Lakkaraju, Leskovec, Ludwig, and
  Mullainathan}{Kleinberg et~al\mbox{.}}{2017a}]%
        {KleinbergLLLM17}
\bibfield{author}{\bibinfo{person}{Jon Kleinberg}, \bibinfo{person}{Himabindu
  Lakkaraju}, \bibinfo{person}{Jure Leskovec}, \bibinfo{person}{Jens Ludwig},
  {and} \bibinfo{person}{Sendhil Mullainathan}.}
  \bibinfo{year}{2017}\natexlab{a}.
\newblock \bibinfo{booktitle}{{\em Human Decisions and Machine Predictions}}.
\newblock \bibinfo{type}{Working Paper} 23180. \bibinfo{institution}{National
  Bureau of Economic Research}.
\newblock
\showDOI{%
\url{https://doi.org/10.3386/w23180}}


\bibitem[\protect\citeauthoryear{Kleinberg, Ludwig, Mullainathan, and
  Obermeyer}{Kleinberg et~al\mbox{.}}{2015}]%
        {KleinbergLMO15}
\bibfield{author}{\bibinfo{person}{Jon Kleinberg}, \bibinfo{person}{Jens
  Ludwig}, \bibinfo{person}{Sendhil Mullainathan}, {and} \bibinfo{person}{Ziad
  Obermeyer}.} \bibinfo{year}{2015}\natexlab{}.
\newblock \showarticletitle{Prediction Policy Problems}.
\newblock \bibinfo{journal}{{\em American Economic Review\/}}
  \bibinfo{volume}{105}, \bibinfo{number}{5} (\bibinfo{date}{May}
  \bibinfo{year}{2015}), \bibinfo{pages}{491--95}.
\newblock
\showDOI{%
\url{https://doi.org/10.1257/aer.p20151023}}


\bibitem[\protect\citeauthoryear{Kleinberg, Mullainathan, and
  Raghavan}{Kleinberg et~al\mbox{.}}{2017b}]%
        {KleinbergMR17}
\bibfield{author}{\bibinfo{person}{Jon Kleinberg}, \bibinfo{person}{Sendhil
  Mullainathan}, {and} \bibinfo{person}{Manish Raghavan}.}
  \bibinfo{year}{2017}\natexlab{b}.
\newblock \showarticletitle{Inherent Trade-Offs in the Fair Determination of
  Risk Scores}. In \bibinfo{booktitle}{{\em Proceedings of the 8th Innovations
  in Theoretical Computer Science Conference}} {\em
  (\bibinfo{series}{ITCS~'17})}.
\newblock


\bibitem[\protect\citeauthoryear{Leslie and Pope}{Leslie and Pope}{2016}]%
        {leslie2016unintended}
\bibfield{author}{\bibinfo{person}{Emily Leslie} {and} \bibinfo{person}{Nolan~G
  Pope}.} \bibinfo{year}{2016}\natexlab{}.
\newblock \showarticletitle{The Unintended Impact of Pretrial Detention on Case
  Outcomes: Evidence from NYC Arraignments}.
\newblock  (\bibinfo{year}{2016}).
\newblock


\bibitem[\protect\citeauthoryear{Lum and Yang}{Lum and Yang}{2005}]%
        {lum2005evaluation}
\bibfield{author}{\bibinfo{person}{Cynthia Lum} {and} \bibinfo{person}{Sue-Ming
  Yang}.} \bibinfo{year}{2005}\natexlab{}.
\newblock \showarticletitle{Why do evaluation researchers in crime and justice
  choose non-experimental methods?}
\newblock \bibinfo{journal}{{\em Journal of Experimental Criminology\/}}
  \bibinfo{volume}{1}, \bibinfo{number}{2} (\bibinfo{year}{2005}),
  \bibinfo{pages}{191--213}.
\newblock


\bibitem[\protect\citeauthoryear{Maurutto and Hannah-Moffat}{Maurutto and
  Hannah-Moffat}{2006}]%
        {MauruttoHM06}
\bibfield{author}{\bibinfo{person}{Paula Maurutto} {and} \bibinfo{person}{Kelly
  Hannah-Moffat}.} \bibinfo{year}{2006}\natexlab{}.
\newblock \showarticletitle{Assembling Risk and the Restructuring of Penal
  Control}.
\newblock \bibinfo{journal}{{\em The British Journal of Criminology\/}}
  \bibinfo{volume}{46}, \bibinfo{number}{3} (\bibinfo{year}{2006}),
  \bibinfo{pages}{438--454}.
\newblock
\showDOI{%
\url{https://doi.org/10.1093/bjc/azi073}}


\bibitem[\protect\citeauthoryear{Mitchell, Cochran, Mears, and Bales}{Mitchell
  et~al\mbox{.}}{2017}]%
        {MitchellCMB17}
\bibfield{author}{\bibinfo{person}{Ojmarrh Mitchell},
  \bibinfo{person}{Joshua~C. Cochran}, \bibinfo{person}{Daniel~P. Mears}, {and}
  \bibinfo{person}{William~D. Bales}.} \bibinfo{year}{2017}\natexlab{}.
\newblock \showarticletitle{Examining Prison Effects on Recidivism: A
  Regression Discontinuity Approach}.
\newblock \bibinfo{journal}{{\em Justice Quarterly\/}} \bibinfo{volume}{34},
  \bibinfo{number}{4} (\bibinfo{year}{2017}), \bibinfo{pages}{571--596}.
\newblock


\bibitem[\protect\citeauthoryear{Monahan and Skeem}{Monahan and Skeem}{2016}]%
        {MonahanS16}
\bibfield{author}{\bibinfo{person}{John Monahan} {and}
  \bibinfo{person}{Jennifer~L. Skeem}.} \bibinfo{year}{2016}\natexlab{}.
\newblock \showarticletitle{Risk Assessment in Criminal Sentencing}.
\newblock \bibinfo{journal}{{\em Annual Review of Clinical Psychology\/}}
  \bibinfo{volume}{12}, \bibinfo{number}{1} (\bibinfo{year}{2016}),
  \bibinfo{pages}{489--513}.
\newblock
\showDOI{%
\url{https://doi.org/10.1146/annurev-clinpsy-021815-092945}}


\bibitem[\protect\citeauthoryear{Paluck}{Paluck}{2010}]%
        {Paluck10}
\bibfield{author}{\bibinfo{person}{Elizabeth~Levy Paluck}.}
  \bibinfo{year}{2010}\natexlab{}.
\newblock \showarticletitle{The Promising Integration of Qualitative Methods
  and Field Experiments}.
\newblock \bibinfo{journal}{{\em The Annals of the American Academy of
  Political and Social Science\/}}  \bibinfo{volume}{628}
  (\bibinfo{year}{2010}), \bibinfo{pages}{59--71}.
\newblock


\bibitem[\protect\citeauthoryear{Prins and Reich}{Prins and Reich}{2017}]%
        {prins2017can}
\bibfield{author}{\bibinfo{person}{Seth~J Prins} {and} \bibinfo{person}{Adam
  Reich}.} \bibinfo{year}{2017}\natexlab{}.
\newblock \showarticletitle{Can we avoid reductionism in risk reduction?}
\newblock \bibinfo{journal}{{\em Theoretical Criminology\/}}
  (\bibinfo{year}{2017}), \bibinfo{pages}{1362480617707948}.
\newblock


\bibitem[\protect\citeauthoryear{Reisig, Holtfreter, and Morash}{Reisig
  et~al\mbox{.}}{2006}]%
        {ReisigHM06}
\bibfield{author}{\bibinfo{person}{Michael~D. Reisig}, \bibinfo{person}{Kristy
  Holtfreter}, {and} \bibinfo{person}{Merry Morash}.}
  \bibinfo{year}{2006}\natexlab{}.
\newblock \showarticletitle{Assessing Recidivism Risk Across Female Pathways to
  Crime}.
\newblock \bibinfo{journal}{{\em Justice Quarterly\/}} \bibinfo{volume}{23},
  \bibinfo{number}{3} (\bibinfo{year}{2006}), \bibinfo{pages}{384--405}.
\newblock
\showDOI{%
\url{https://doi.org/10.1080/07418820600869152}}


\bibitem[\protect\citeauthoryear{{Roggensack, C.J.} and {Abrahamson,
  J.}}{{Roggensack, C.J.} and {Abrahamson, J.}}{2016}]%
        {Loomis15}
\bibfield{author}{\bibinfo{person}{{Roggensack, C.J.}} {and}
  \bibinfo{person}{{Abrahamson, J.}}} \bibinfo{year}{2016}\natexlab{}.
\newblock \bibinfo{title}{State of Wisconsin v. Eric L. Loomis}.
\newblock   (\bibinfo{date}{July} \bibinfo{year}{2016}).
\newblock
\showURL{%
\url{https://www.wicourts.gov/sc/opinion/DisplayDocument.pdf?content=pdf&seqNo=171690}}


\bibitem[\protect\citeauthoryear{Rubin}{Rubin}{2006}]%
        {rubin2006matched}
\bibfield{author}{\bibinfo{person}{Donald~B Rubin}.}
  \bibinfo{year}{2006}\natexlab{}.
\newblock \bibinfo{booktitle}{{\em Matched sampling for causal effects}}.
\newblock \bibinfo{publisher}{Cambridge University Press}.
\newblock


\bibitem[\protect\citeauthoryear{Schlesinger}{Schlesinger}{2011}]%
        {Schlesinger11}
\bibfield{author}{\bibinfo{person}{Traci Schlesinger}.}
  \bibinfo{year}{2011}\natexlab{}.
\newblock \showarticletitle{The Failure of Race Neutral Policies: How Mandatory
  Terms and Sentencing Enhancements Contribute to Mass Racialized
  Incarceration}.
\newblock \bibinfo{journal}{{\em Crime \& Delinquency\/}} \bibinfo{volume}{57},
  \bibinfo{number}{1} (\bibinfo{year}{2011}), \bibinfo{pages}{56--81}.
\newblock
\showDOI{%
\url{https://doi.org/10.1177/0011128708323629}}


\bibitem[\protect\citeauthoryear{Selbst}{Selbst}{2016}]%
        {Selbst16}
\bibfield{author}{\bibinfo{person}{Andrew~D. Selbst}.}
  \bibinfo{year}{2016}\natexlab{}.
\newblock \showarticletitle{Disparate Impact in Big Data Policing}.
\newblock \bibinfo{journal}{{\em Georgia Law Review\/}} (\bibinfo{year}{2016}).
\newblock
\newblock
\shownote{Forthcoming; SSRN ID: 2819182.}


\bibitem[\protect\citeauthoryear{Sherman}{Sherman}{2006}]%
        {sherman2002}
\bibfield{author}{\bibinfo{person}{Lawrence Sherman}.}
  \bibinfo{year}{2006}\natexlab{}.
\newblock \bibinfo{booktitle}{{\em Evidence-based crime prevention}}.
\newblock \bibinfo{publisher}{Routledge}, \bibinfo{address}{London New York}.
\newblock
\showISBNx{978-0415401029}


\bibitem[\protect\citeauthoryear{Simpson}{Simpson}{1989}]%
        {Simpson89}
\bibfield{author}{\bibinfo{person}{Sally~S. Simpson}.}
  \bibinfo{year}{1989}\natexlab{}.
\newblock \showarticletitle{Feminist Theory, Crime, and Justice}.
\newblock \bibinfo{journal}{{\em Criminology\/}} \bibinfo{volume}{27},
  \bibinfo{number}{4} (\bibinfo{year}{1989}), \bibinfo{pages}{605--632}.
\newblock
\showISSN{1745-9125}
\showDOI{%
\url{https://doi.org/10.1111/j.1745-9125.1989.tb01048.x}}


\bibitem[\protect\citeauthoryear{Siwicki}{Siwicki}{2017}]%
        {Siwicki17}
\bibfield{author}{\bibinfo{person}{Bill Siwicki}.}
  \bibinfo{year}{2017}\natexlab{}.
\newblock \bibinfo{title}{Machine learning 101: The healthcare opportunities
  are endless}.
\newblock   (\bibinfo{year}{2017}).
\newblock
\showURL{%
\url{http://www.healthcareitnews.com/news/machine-learning-101-healthcare-opportunities-are-endless}}


\bibitem[\protect\citeauthoryear{Skeem, Monahan, and Lowenkamp}{Skeem
  et~al\mbox{.}}{2016}]%
        {SkeemML16}
\bibfield{author}{\bibinfo{person}{Jennifer~L. Skeem}, \bibinfo{person}{John
  Monahan}, {and} \bibinfo{person}{Christopher~T. Lowenkamp}.}
  \bibinfo{year}{2016}\natexlab{}.
\newblock \bibinfo{booktitle}{{\em Gender, Risk Assessment, and Sanctioning:
  The Cost of Treating Women Like Men}}.
\newblock \bibinfo{type}{Working Paper}~10. \bibinfo{institution}{Virginia
  Public Law and Legal Theory Research Paper Series}.
\newblock


\bibitem[\protect\citeauthoryear{Starr}{Starr}{2014}]%
        {Starr14}
\bibfield{author}{\bibinfo{person}{Sonja~B. Starr}.}
  \bibinfo{year}{2014}\natexlab{}.
\newblock \showarticletitle{Evidence-Based Sentencing and the Scientific
  Rationalization of Discrimination}.
\newblock \bibinfo{journal}{{\em Stanford Law Review\/}}  \bibinfo{volume}{66}
  (\bibinfo{year}{2014}).
\newblock
Issue 4.


\bibitem[\protect\citeauthoryear{Stevenson}{Stevenson}{2016}]%
        {stevenson2016distortion}
\bibfield{author}{\bibinfo{person}{Megan Stevenson}.}
  \bibinfo{year}{2016}\natexlab{}.
\newblock \showarticletitle{Distortion of justice: How the inability to pay
  bail affects case outcomes}.
\newblock  (\bibinfo{year}{2016}).
\newblock


\bibitem[\protect\citeauthoryear{Tashea}{Tashea}{2017}]%
        {Tashea17}
\bibfield{author}{\bibinfo{person}{Jason Tashea}.}
  \bibinfo{year}{2017}\natexlab{}.
\newblock \bibinfo{title}{Courts are using AI to sentence criminals. That must
  stop now}.
\newblock   (\bibinfo{year}{2017}).
\newblock
\showURL{%
\url{https://www.wired.com/2017/04/courts-using-ai-sentence-criminals-must-stop-now/}}


\bibitem[\protect\citeauthoryear{Van~Cleve and Mayes}{Van~Cleve and Mayes}{[n.
  d.]}]%
        {VanCleveM15}
\bibfield{author}{\bibinfo{person}{Nicole~Gonzalez Van~Cleve} {and}
  \bibinfo{person}{Lauren Mayes}.} \bibinfo{year}{[n. d.]}\natexlab{}.
\newblock \showarticletitle{Criminal Justice Through ``Colorblind'' Lenses: A
  Call to Examine the Mutual Constitution of Race and Criminal Justice}.
\newblock \bibinfo{journal}{{\em Law \& Social Inquiry\/}}
  \bibinfo{volume}{40}, \bibinfo{number}{2} (\bibinfo{year}{[n. d.]}).
\newblock
\showISSN{1747-4469}


\bibitem[\protect\citeauthoryear{Vellido, Mart\'{i}n-Guerrero, and
  Lisboa}{Vellido et~al\mbox{.}}{2012}]%
        {VellidoMGL12}
\bibfield{author}{\bibinfo{person}{Alfredo Vellido}, \bibinfo{person}{Jos\'{e}
  Mart\'{i}n-Guerrero}, {and} \bibinfo{person}{Paulo~J.G. Lisboa}.}
  \bibinfo{year}{2012}\natexlab{}.
\newblock \showarticletitle{Making machine learning models interpretable}.
\newblock   \bibinfo{volume}{12} (\bibinfo{date}{01} \bibinfo{year}{2012}),
  \bibinfo{pages}{163--172}.
\newblock


\bibitem[\protect\citeauthoryear{Wang}{Wang}{2017}]%
        {Wang17}
\bibfield{author}{\bibinfo{person}{Tammy Wang}.}
  \bibinfo{year}{2017}\natexlab{}.
\newblock \bibinfo{title}{How Machine Learning Will Shape the Future of
  Hiring}.
\newblock   (\bibinfo{year}{2017}).
\newblock
\showURL{%
\url{https://www.linkedin.com/pulse/how-machine-learning-shape-future-hiring-tammy-wang}}


\bibitem[\protect\citeauthoryear{Weisburd}{Weisburd}{2000}]%
        {weisburd2000randomized}
\bibfield{author}{\bibinfo{person}{David Weisburd}.}
  \bibinfo{year}{2000}\natexlab{}.
\newblock \showarticletitle{Randomized experiments in criminal justice policy:
  Prospects and problems}.
\newblock \bibinfo{journal}{{\em NCCD news\/}} \bibinfo{volume}{46},
  \bibinfo{number}{2} (\bibinfo{year}{2000}), \bibinfo{pages}{181--193}.
\newblock


\bibitem[\protect\citeauthoryear{Wexler}{Wexler}{2017}]%
        {Wexler17}
\bibfield{author}{\bibinfo{person}{Rebecca Wexler}.}
  \bibinfo{year}{2017}\natexlab{}.
\newblock \bibinfo{title}{Code of Sielnce: How private companies hide flaws in
  the software that governments use to decide who goes to prison and who gets
  out}.
\newblock   (\bibinfo{year}{2017}).
\newblock
\showURL{%
\url{http://washingtonmonthly.com/magazine/junejulyaugust-2017/code-of-silence/}}


\bibitem[\protect\citeauthoryear{Zeng, Ustun, and Rudin}{Zeng
  et~al\mbox{.}}{2017}]%
        {ZengUR17}
\bibfield{author}{\bibinfo{person}{Jiaming Zeng}, \bibinfo{person}{Berk Ustun},
  {and} \bibinfo{person}{Cynthia Rudin}.} \bibinfo{year}{2017}\natexlab{}.
\newblock \showarticletitle{Interpretable classification models for recidivism
  prediction}.
\newblock \bibinfo{journal}{{\em Journal of the Royal Statistical Society:
  Series A (Statistics in Society)\/}} \bibinfo{volume}{180},
  \bibinfo{number}{3} (\bibinfo{year}{2017}), \bibinfo{pages}{689--722}.
\newblock
\showISSN{1467-985X}
\showDOI{%
\url{https://doi.org/10.1111/rssa.12227}}


\bibitem[\protect\citeauthoryear{Zimring}{Zimring}{2006}]%
        {Zimring06}
\bibfield{author}{\bibinfo{person}{Franklin~E. Zimring}.}
  \bibinfo{year}{2006}\natexlab{}.
\newblock \bibinfo{booktitle}{{\em The Great American Crime Decline}}.
\newblock \bibinfo{publisher}{Oxford University Press}. 258 pages.
\newblock
\showDOI{%
\url{https://doi.org/10.1093/acprof:oso/9780195181159.001.0001}}


\end{thebibliography}
